\documentclass[article]{llncs}

\usepackage{graphicx}
\usepackage{booktabs}

\usepackage[accsupp]{axessibility}

\usepackage{hyperref}

\usepackage{orcidlink}

\newcommand{\courier}[1]{\texttt{#1}}

\newcounter{mariette}
\stepcounter{mariette}

\usepackage{adjustbox, float, multirow, subcaption}

\begin{document}

\title{Transfer learning RGB models to hyperspectral images with trainable tensor decompositions} 

\titlerunning{Transfer learning RGB models to hyperspectral images}

\author{Mariette Sch\"onfeld \inst{1,2}\orcidlink{0009-0000-3335-6538} \and
Laurens Devos\inst{1,2}\orcidlink{0000-0002-1549-749X} \and
Wannes Meert \inst{1,2}\orcidlink{0000-0001-9560-3872} \and
Hendrik Blockeel\inst{1,2}\orcidlink{0000-0003-0378-3699}}

\authorrunning{M.~Schönfeld et al.}

\institute{
KU Leuven, Dept. of Computer Science, B-3000 Leuven, Belgium 
\email{\{mariette.schoenfeld,laurens.devos,wannes.meert,hendrik.blockeel\}@kuleuven.be}
\url{https://wms.cs.kuleuven.be/cs/english}
\and 
Leuven.AI - KU Leuven Institute for AI, B-3000 Leuven, Belgium
\url{https://ai.kuleuven.be/}
}

\maketitle

\begin{abstract}
Transfer learning makes it possible to use large vision networks on a variety of domains, by specializing their models' general filters to new tasks. However, these networks assume the input images to have 3 input channels, making them incompatible with multi- or hyperspectral images. Current approaches that mitigate this incompatibility sacrifice information in either the image, or the model. This work proposes a novel approach that preserves the image and spatial information present in the model by using partially trainable tensor decompositions. We create such decompositions of pretrained convolutional filters, separating the filters into spatial and spectral components. The spectral components are then replaced with trainable components of higher channel dimensionality. This creates hyperspectral filters that can specialize to new datasets, while retaining the spatial patterns of the original filter. Experiments on a variety of hyperspectral datasets show that our approach is more accurate and robust than other hyperspectral transfer learning methods. 

\end{abstract}
\section{Introduction}
Hyperspectral images are images where each pixel is described by the intensities of many wavelengths (possibly hundreds), within or beyond the visible spectrum, rather than the three RGB values typically used. The additional information in this description makes no difference to the human eye, but can be crucial for data analysis. Hyperspectral images have proven valuable for a variety of complex vision tasks, including satellite imagery \cite{zhu2017deep}, medical imaging \cite{lu2014medical}, and horticulture \cite{lu2020hyperspectral}.

Hyperspectral image processing can roughly be categorized into remote sensing and near-range image processing.  In remote sensing \cite{zhu2017deep}, individual pixels are typically classified based on their own spectrum, and downstream tasks such as segmentation make use of that information.  Near-range image processing \cite{lu2014medical,lu2020hyperspectral} is more similar to classical vision in the sense that spatial patterns (such as those learned by convolutional neural networks) are often relevant. In this paper, we mostly focus on this second type. A challenge in training models on hyperspectral images is the acquisition of large amounts of hyperspectral data: this is expensive, time-consuming, and logistically challenging. As a result, datasets are typically small, making hyperspectral imaging an excellent candidate for transfer learning \cite{shao2014transfer}. 

However, transfer learning requires a \textit{backbone model} to transfer knowledge from, and a hyperspectral backbone is unlikely to be developed soon, for two reasons. First, the potential of transfer learning greatly depends on the variety and amount of data that the backbone model has seen. The most versatile and accurate backbones owe their success to their universally applicable filters, a result of training extensively on large, diverse datasets like ImageNet \cite{deng2009imagenet}. Because the amount and variety of hyperspectral data is significantly lower than for RGB images, creating backbone models of comparable quality for hyperspectral data is difficult. Second, the lack of a common data format makes the existence of a hyperspectral backbone impractical. Hyperspectral datasets vary in what wavelengths they use (and how many), whereas RGB datasets always use the same three wavelengths. Applying transfer learning in hyperspectral imaging therefore requires bridging varying input dimensionalities. For example, building a classification model for 50-dimensional inputs by transferring a 3-dimensional RGB backbone model necessitates addressing the input-channel gap.

Several solutions have been proposed for solving this mismatch between hyperspectral data and current backbones. A lightweight solution is to perform dimensionality reduction on the \textit{spectral} (i.e. channel or color) dimension of the image, in order to create a three-channel extract \cite{giri2022spatial,laprade2024hyperleaf2024}. This sacrifices information in the image, but preserves the backbone. A different approach is to replace the first convolutional layer of the backbone by one with more input channels, while keeping the rest of the network structure intact \cite{varga2021measuring}; during fine-tuning, the new layer is then trained from scratch. This approach allows the new model to use information across all channels in the image, but has the downside that the backbone's first-layer filters are sacrificed. Furthermore, this also increases the number of trainable parameters and therefore the risk of overfitting to spurious correlations grows.

This paper proposes a novel approach for tackling the above-mentioned mismatch. It is known that weights of backbone models can be compressed with minimal performance loss using tensor decompositions \cite{denton2014exploiting} because of the relative simplicity of the learned filters \cite{zeiler2014visualizing}. These techniques are typically used with the goal of model compression, but tensor decompositions have another advantage: by design, they decouple spatial and spectral patterns \cite{kolda2009tensor}. This paper builds on this property to facilitate transfer learning from 3-channel RGB backbones for hyperspectral tasks. We accomplish this by decomposing the convolutional filters in the first layer, separating the filters into their spatial and spectral components. Each 3-channel spectral component is then replaced by a higher-dimensional, hyperspectral one. Decompressing now provides a  filter that can be convolved with the hyperspectral image, and contains the spatial patterns of the original filter. The new components can then be trained in a standard transfer learning pipeline. This solves the dimensionality mismatch between the hyperspectral image and the backbone model, preserves both the model and the image, and enables a traditional transfer learning pipeline for hyperspectral images with a minimal number of trainable parameters.

We first provide background in section \ref{sec:background} on transfer learning and tensor decompositions. Then, we introduce our method in section \ref{section:method}. Section \ref{sec:relwork} covers related work with regards to transfer learning in hyperspectral imaging and tensor decompositions in neural networks. Finally, we show through experiments in section \ref{sec:experiments} how our method is more accurate and more robust against overfitting than other hyperspectral transfer learning approaches.

\section{Background} \label{sec:background}


\subsection{Transfer learning}
Transfer Learning is a collection of techniques that involve the transfer of previously gained knowledge to a new task \cite{shao2014transfer}.
This can be as simple as starting the learning process from pretrained weights instead of random initializations with the goal accelerating convergence.
In vision contexts however, it can be difficult to prevent overfitting in low-data scenarios and so it is oftentimes recommended \cite{torralba2024foundations} to freeze the feature extraction (i.e. convolutional) stage of a network, and only train the classification layers. This difference leads to a distinction we make in this work: a \textit{pretrained} model versus a \textit{backbone} model. In hyperspectral images, it is common to pretrain a model on related datasets \cite{windrim2018pretraining}, but this is unlikely to result in a model so versatile that transferring from a remote sensing task to a medical task can be done by just retraining the classification layer. With a backbone model, we specifically refer to a model that has been pretrained on large volumes of diverse data (like ImageNet), and can be applied to a variety of vision tasks with minimal retraining. Such backbone models are widely available for regular vision, but do not exist for hyperspectral vision. 

One interesting observation of such backbone models is that the filters in their first layer tend to exhibit colored, elliptical shapes, which cause edge-detection-like behavior \cite{zeiler2014visualizing}. Such filters contain both \textit{spatial} and \textit{spectral} information. With \textit{spatial} we mean the size, orientation, and relative location of the shapes, and with \textit{spectral} we mean their colors.  Although this behavior is not explicitly induced in the network, the fact that it seems to happen for all backbone models gives the impression that it is important and should be preserved for robust and accurate vision. 

\subsection{Tensor decompositions for convolutional filters}
Tensor decompositions are a collection of techniques for approximating complex tensors with smaller, simpler tensors \cite{kolda2009tensor}. Many decompositions are higher-order generalizations of matrix decompositions, where \textit{order} refers to the number of indices in the array. A vector is a first order tensor, a matrix a second order, a cube a third order tensor, e.t.c. Many matrix operations have tensor generalizations, like outer products, or tensor multiplication. Although matrix decompositions can be applied to tensors by \textit{unfolding} them to a matrix, tensor decompositions can preserve the higher-order structure of the tensor by using components for each order. For example, an image can be seen as a third-order tensor with one channel dimension and two spatial dimensions, and a decomposition could involve three projection matrices, each acting on its own order. 

Matrix decompositions were already of interest to deep learning researchers as a tool for network compression, but tensor decompositions were more appealing for two reasons \cite{panagakis2021tensor}: they offer the opportunity to rewrite the convolutional operator directly over the components of the decomposition instead of having to decompress the filters. This makes tensor decomposed layers not just more efficient for storage, but also for inferencing \cite{jaderberg2014speeding,kim2015compression,lebedev2015speeding}. Second, it has been observed that the aforementioned elliptical shapes in backbone filters have high similarity to either low-rank or multilinear patterns \cite{denton2014exploiting}, the mathematical structures that tensor decomposition approximate tensors with. As there are multiple types of decompositions and ways to apply them to a tensor, there is also not a single universal method to use them for model compression. 
\section{Filter decompositions for hyperspectral transfer learning}\label{section:method}
The previous sections can be summarized as follows: hyperspectral vision could benefit from transfer learning. This is hindered by the lack of hyperspectral backbone models and the dimensionality mismatch between hyperspectral images and 3-channel backbone models. Such backbone models typically exhibit colored, elliptical shapes, which should ideally be preserved. These filters lend themselves well to tensor decomposed convolutions, which split the filters into spatial and spectral components. We will use this split to solve the mismatch between backbone models and hyperspectral images. 

\begin{figure}[t]
    \centering
    \includegraphics[width=\linewidth]{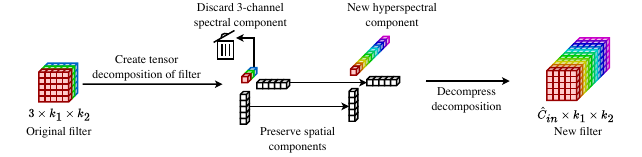}
    \caption{Illustration of our general approach. Values for the new hyperspectral component can be learned in a transfer learning pipeline, while keeping the spatial components frozen.}
    \label{fig:tensor_decomposition}
\end{figure}

Consider a backbone model. Typically the first layer is a 2D convolution with a fourth order weight tensor $W^1$ of shape $C_{out}\times C_{in} \times k_1 \times k_2$ with $C_{out}$ the number of filters or output channels, $C_{in}=3$ the input channels, and $k_1$ and $k_2$ the height and width of the filters. An input image needs to match the number of input channels $C_{in}$ and so $W^1$ cannot be convolved with a hyperspectral image $X$ of shape $\hat{C}_{in}\times H \times W$ because $\hat{C}_{in}>3$. We approximate the filters in $W^1$ with tensor decompositions, providing us with spatial and 3-channel spectral components. We can then replace the spectral components with larger ones that match $\hat{C}_{in}$. Decompressing the decomposition would give $C_{out}$ filters of shape $\hat{C}_{in} \times k_1 \times k_2$, which contain the spatial patterns of the original filter but can be convolved with $X$. A transfer learning pipeline can then be used for learning informative values for the spectral components. 
We discuss this approach (visualized in figure \ref{fig:tensor_decomposition}) for CP and Tucker decompositions, the two most popular decompositions \cite{kolda2009tensor}. 

\begin{figure}[b]
    \centering
    \includegraphics[width=\linewidth]{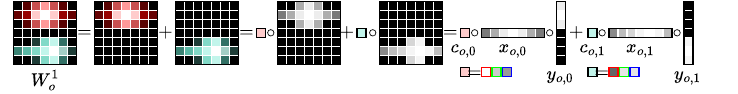}
    \caption{Illustration of how a filter is decomposed into its spectral ($c_o^i$) and spatial ($x_o^i$ and $y_o^i$) components with a CP decomposition.}
    \label{fig:cp_decomposition}
\end{figure}
\subsection{CP decompositions}
CP (Canonic Polyadic) decompositions approximate tensors as a finite sum of \textit{rank-one tensors}. A rank-one tensor is a tensor that can be written as the outer product of vectors (illustrated in figure \ref{fig:cp_decomposition}). This makes a second order rank-one tensor the same as a separable filter \cite{rigamonti2013learning}, and a third order rank-one tensor can be seen as a colored separable filter with two spatial vectors, and one spectral vector indicating the color. Changing the values in this third vector only changes the colors in the filter, but not the shapes. The \textit{rank} of a decomposition then refers to how many of such filters can be part of the summation. The illustration of this decomposition in figure \ref{fig:cp_decomposition} demonstrates how changing the values in the spectral vectors only affects the colors in the filter, not the shapes. We approximate a filter $W_o^1$ with CP decomposition of rank $R$ as:
\begin{equation}
    W_o^1 \approx \overline{W_o^1} = \sum_{r=0}^R c_{o,r} \circ x_{o,r} \circ y_{o,r},\  0 \leq o < C_{out}
    \label{eq:cp_filters}
\end{equation}
Where $c_{o,r}$ is the $r$'th spectral component of shape $3$, $x_{o,r}$ the $r$'th horizontal component of shape $k_1$, and $y_{o,r}$ the $r$'th vertical components of shape $k_2$. The spectral components $c_{o,r}$ are then replaced with new, trainable components $\hat{c}_{o,r}$ of size $\hat{C}_{in}$.

\subsection{Tucker decompositions}
In a Tucker decomposition, the original tensor is projected into a lower dimensional core tensor through matrix multiplications over every order. An intuitive example of this idea is pressing on a foam cube with three pairs of hands, each pair pressing on two opposite sides. Every pair can exert different amounts of force, thus forcing the cube into a smaller, irregular, cuboid. In this example, the compressed foam cube symbolizes the core tensor, and the three pairs of hands compressing the foam cube signify the component matrices. This `multi-directional' compression also makes it possible to do a \textit{partial} Tucker decomposition: compressing only a subset of the orders (metaphorically: less than 3 pairs of hands pressing on the cube). Formally, we approximate $W_o^1$ through a partial Tucker decomposition over only the spectral dimension of rank $R$ as:
\begin{equation}
    W_o^1 \approx \overline{W_o^1} = V_o \times_1 A_{o},\  0 \leq o < C_{out}
    \label{eq:tucker_filters}
\end{equation}
where $\times_1$ is the 1-mode product \cite{kolda2009tensor}, $V_o$ the \textit{core tensor} of shape $R\times k_1 \times k_2$ and $A_o$ the spectral component of shape $3\times R$. We then replace $A_o$ with a new component $\hat{A}_o$ of shape $C_{in}\times R$, again creating a filter of shape $\hat{C}_{in} \times k_1 \times k_2$.

Both CP and Tucker decompositions carry specific advantages and disadvantages, so choosing which one to use depends on the task at hand. Both decompositions can be calculated numerically by applying either Alternating Least Squares or Singular Value Decomposition iteratively and minimizing an error between the original tensor and the result of the decomposition \cite{kolda2009tensor}. Computing the decompositions needs to happen only once, and requires negligible resources due to the small size of the individual filters.
\section{Related work}\label{sec:relwork}
\subsection{Transfer learning for hyperspectral images}
Hyperspectral transfer learning works can be roughly divided into two categories: they either create compatibility with 3-channel backbone models, or pretrain a new model on multiple hyperspectral datasets.

\subsubsection{Using backbone models}
Compatibility between 3-channel backbone models and hypersectral images can be created by either adapting the image, or the model. Solving the mismatch on the image side involves creating a 3-channel extract of the hyperspectral image. This can be done through traditional dimensionality reduction techniques, like PCA \cite{giri2022spatial}, or channel selection \cite{corley2024revisiting}. It is also possible to include the dimensionality reduction into the network, for example with convolutions \cite{laprade2024hyperleaf2024}. However, such 3-channel extracts have been shown to not preserve all the relevant information of the hyperspectral image \cite{varga2023wavelength}. 

Tackling the dimensionality mismatch on the model side preserves the image, but does involve a sacrifice of information on the backbone's side. For example by replacing the first layer with one that has more input channels \cite{varga2021measuring}. The new weights can then be initialized by expanding the previously learned filters, either by upsampling or stacking the same filter multiple times \cite{windrim2018pretraining}. If one does not retrain, such initialization methods are functionally identical to channel dimensionality reduction. Retraining is likely to destroy the original filters though, adds a significant amount of trainable parameters, and increases the risk of overfitting to spurious correlations. 

All hyperspectral transfer learning techniques can benefit from spectral attention methods like Squeeze-Excitation Blocks \cite{hu2018squeeze}, Convolutional Block Attention Modules \cite{woo2018cbam} or Wavelength-aware Convolutions \cite{varga2023wavelength}. Attention can help highlight the most informative channels, but it does not solve the dimensionality mismatch between the image and the model itself and still requires compressing the image to 3 channels, or adapting the first layer of the model. 

\subsubsection{Pretraining}
While we focus on adapting backbone models in this work, a significant portion of hyperspectral transfer learning research focuses on transferring between closely related hyperspectral datasets \cite{wang2022empirical}. For example by pretraining a model on a larger remote sensing dataset before fine-tuning to a smaller one \cite{wang2022empirical}, merging smaller datasets with similar classes \cite{windrim2018pretraining}, or self-supervised pre-training on unlabelled hyperspectral data \cite{varga2021measuring}. A large theme in this research is how to combine hyperspectral datasets with different amounts of channels. A straightforward solution to this problem is to either down- or upsample the images in the spectral dimension \cite{corley2024revisiting}. Instead of addressing the mismatch in the channels, one can also convert the spectral dimension to a third spatial dimension and use 3D convolutions \cite{zhang2019hyperspectral}. Differing amounts of channels then results in differently sized feature maps. However, 3D convolutions have also been shown to add significant computational load, and do not beat 2D backbone models~\cite{laprade2024hyperleaf2024}. 

\subsection{Decomposing backbone models}
The observation that backbone filters tend to decompose well has led to a variety of research, the majority of which focuses on model compression. Already in 2014, low-rank filter decompositions were shown to compress pretrained networks with minimal sacrifices in accuracy \cite{denton2014exploiting,jaderberg2014speeding}. These works also demonstrated the possibility of inferencing over the components of the decomposition instead of decompressing the filter, reducing the amount of operations. More recently, tensor decompositions \cite{kolda2009tensor} have been explored for filter decompositions, either with CP decompositions \cite{lebedev2015speeding}, or Tucker decompositions \cite{kim2015compression}. In contrast with low-rank decompositions, tensor decompositions can preserve the three-dimensional structure of individual convolutional filters, or even find inter-filter correlations by decomposing the four-dimensional filter banks. 

Low-rank decompositions have also become popular for Natural Language Processing \cite{hu2022lora}. However, they are used mostly for the sake of efficient weight updates for fine-tuning instead of weight compression. Furthermore, low-rank decompositions can also be used to combine models. For example, Low-Rank Mixtures decompose multiple models, mix-and-match their components, and create new models that carry capabilities of all models \cite{wu2024mixture}. Although low-rank adaptation does carry similarity to filter decompositions, the techniques are not trivially interchangeable between language and vision models. 

\subsection{Separable convolutions}
Tensor decomposed filters have a strong relationship with \textit{separable convolutions} \cite{chollet2017xception}. Separable convolutions define their operator with the same operations used in tensor decompositions, and so they can be used for implementing tensor-decomposed layers by assigning the calculated components to the weights \cite{panagakis2021tensor}. The other way around, using the weights of a separable convolution as components for a tensor decomposition would always result in a filter that exhibits low-rank or multilinear patterns. The value of separable convolutions has already been noted for hyperspectral imaging \cite{varga2021measuring,varga2023self}, but combining them with tensor decompositions of backbone filters for this application is to our knowledge novel. In terms of separable convolutions, our approach can be summarized as follows: we replace the first layer of the backbone with a mixture of pointwise and spatially separable convolutions. Tensor decompositions of backbone filters are used to initialize the spatially separable convolutions. The pointwise convolutions are then trained on hyperspectral data, while the spatially separable ones are frozen. 
\section{Experiments}\label{sec:experiments}
We now show the efficacy of our method through empirical experiments over multiple hyperspectral datasets and compare our partially trainable filter decompositions to other hyperspectral transfer learning approaches. With these experiments we aim to answer the following questions: 
\begin{enumerate}
    \item How does our approach compare to other hyperspectral transfer learning approaches? 
    \item Are those conclusion stable when using a different backbone?
    \item How does the choice of decomposition impact results? 
\end{enumerate}
\subsection{Datasets}
\subsubsection{Remote sensing}
We consider three popular remote sensing datasets \cite{indianpines}: `Botswana', `Indian Pines', `Kennedy Space Center', which consist of satellite imagery covering several square kilometers. Every dataset comes with pixel-wise ground truth segmentation, distinguishing classes like `asphalt' or `celery'. Remote sensing datasets typically contain a single image, so we use a method from the remote sensing literature that splits the image into smaller tiles \cite{giri2022spatial}. We split the image into $11 \times 11$ tiles with a stride of 3 and resize them to $32\times 32$, with the center pixel defining the label. We calculate a channel-wise mean-standard deviation shift over the training tiles that we also apply over the test tiles. 

\begin{table}[t]
\centering
\caption{Details on the datasets.}
\label{table:datasets}
\begin{adjustbox}{max width=\linewidth}
\begin{tabular}{llll}
\toprule
\textbf{Dataset} & 
\textbf{Number of tiles/samples} &
\textbf{$\#$Channels}&\textbf{\#Classes} \\
\toprule
Botswana \cite{indianpines}& 340 & 145 & 14\\
Indian Pines \cite{indianpines}& 1146 & 200 & 16 \\  
Kennedy Space Center \cite{indianpines}& 594 & 176 & 13 \\
\midrule
Avocado \cite{varga2021measuring} &  190 & 224 & 3 \\
Grape leaves \cite{ryckewaert2023hyperspectral} & 199 & 194 & 8 \\

\bottomrule
\end{tabular}
\end{adjustbox}
\end{table}
\begin{figure}[H]
    \centering
    \begin{subfigure}{1in}
        \centering
        \includegraphics[width=0.75in]{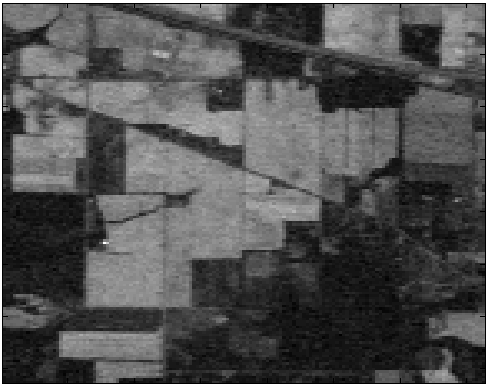} 
        \caption{170th band of the Indian Pines dataset.}
        \label{fig:indian_pines}
    \end{subfigure}
    ~
    \begin{subfigure}{1in}
        \centering
        \includegraphics[width=0.75in]{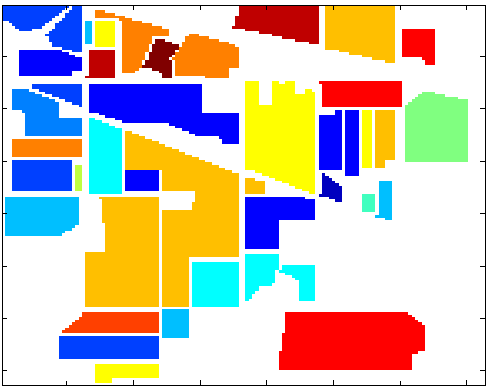} 
        \caption{Ground truth of Indian Pines dataset.}
        \label{fig:indian_pines_dataset}
    \end{subfigure}
    ~
    \begin{subfigure}{1in}
        \centering
        \includegraphics[height=0.75in]{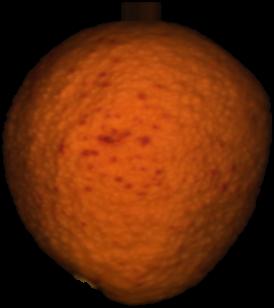} 
        \caption{RGB-extract of avocado image.}
        \label{fig:avocado}
    \end{subfigure}
     ~
    \begin{subfigure}{1in}
        \centering
        \includegraphics[height=0.75in]{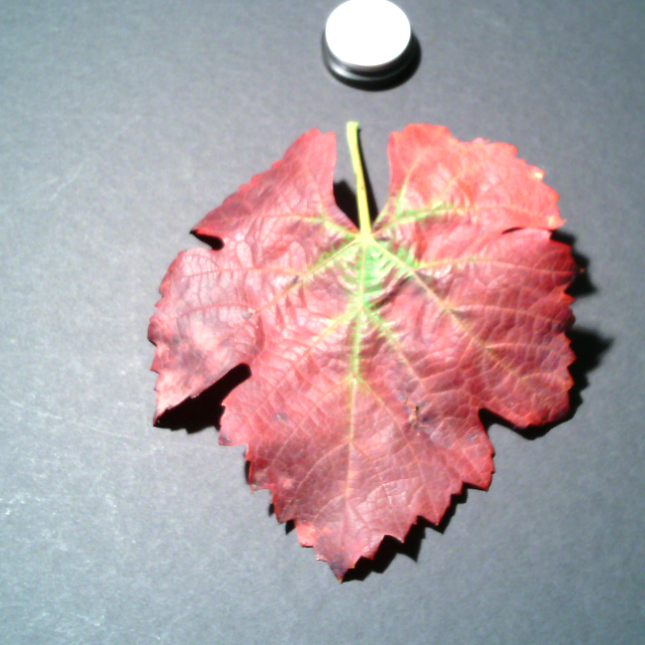} 
        \caption{RGB-extract of grape leaf image.}
        \label{fig:grape_leaf}
    \end{subfigure}
    \caption{}
    \label{fig:images}
\end{figure}

\subsubsection{Horticultural}
We experiment with two horticultural datasets.
The first (`Avocado' \cite{varga2021measuring}) features a ripeness classification scenario. Images range in size and shape, varying between 100 to 300 pixels in height and width. We resize the images to $60 \times 60$ pixels and add 2 pixels of zero padding on all sides, creating $64 \times 64$ images. The second dataset (`Grape leaves' \cite{ryckewaert2023hyperspectral}) features grape leaves affected by various diseases. The images are $512\times 512$ pixels, but most of this area is background. We centercrop the images to $400 \times 400$ pixels, before resizing to $64\times 64$ pixels. We exclude the first and last five spectral channels as recommended by the paper. The dataset includes 12 different diseases of which we exclude four, as there are less than 5 samples of these classes. We compute the same normalization parameters as before for both datasets. 

An overview of the number of samples, number of spectral bands, classes, can be found in table \ref{table:datasets}. Sample images can be found in figure \ref{fig:images}.

\subsection{Models}
\subsubsection{Backbones}
We use pretrained AlexNet \cite{krizhevsky2014one}, DenseNet \cite{huang2017densely}, and ResNet18 \cite{he2016deep}, implemented in the Torchvision library \cite{marcel2010torchvision}. The final feature maps are resized with an adaptive average pooling to a size of $1 \times 1$ for the remote sensing datasets, and $4\times 4$ for the near-range datasets. The feature maps are then flattened and followed by a single linear layer for classification output.

During initial experiments, we noticed that using all blocks of the backbone resulted in unstable gradients in the first layer and found it difficult to acquire stable convergence with any of the models, including our baselines. Using fewer blocks created a stabler gradient, but also larger feature maps. We mitigated this problem by using a smaller image size and using only two blocks of the backbone. 

\subsubsection{Baselines}
We compare our method to two baselines from the literature.
The first, \courier{Reduce}, uses dimensionality reduction to create a 3-channel extract. We follow the approach by Laprade et al. \cite{laprade2024hyperleaf2024} that uses convolutions, but replace their single $3\times 3$ convolution with two $1 \times 1$ convolutions separated by a ReLU activation. The larger $3\times 3$ window size was shown to blur details and using two layers allowed us to adjust the number of neurons to match the number of (trainable) neurons in the filter decompositions.
\begin{figure}[t]
    \centering
    \includegraphics[width=\linewidth]{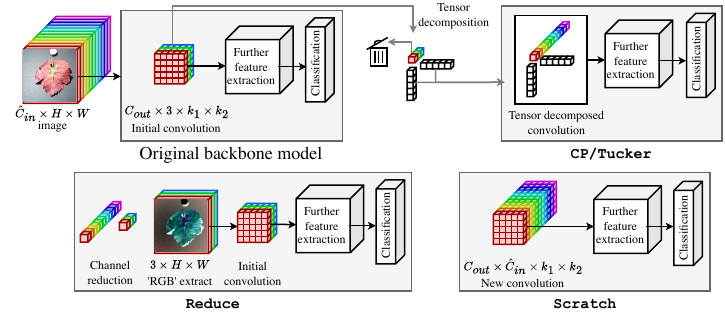}
    \caption{Architecture of the different transfer learning approaches with regards to the backbone model.}
    \label{fig:model_architecture}
\end{figure}
The second baseline, \courier{Scratch}, replaces the first layer of the backbone with one of equal dimensionality, except that it uses $\hat{C}_{in}$ input channels instead of $C_{in}$ \cite{varga2021measuring}. This layer is trained from scratch, offering more flexibility at the risk of overfitting. It uses significantly more trainable parameters ($\frac{k_1 \cdot k_2 \cdot S \cdot O}{R\cdot S\cdot O} = \frac{k_1\cdot k_2}{R}$ times as many), which can only be reduced by decreasing the size of the kernel.  

\subsubsection{Decomposed filters (\courier{CP}, \courier{Tucker})}
We generate decompositions for every filter in the first layer with the TensorLy library \cite{kossaifi2019tensorly}. We implement the tensor decomposed layers with grouped separable convolutions \cite{panagakis2021tensor,chollet2017xception}. \courier{CP} can be implemented with a pointwise convolution, followed by two depthwise convolutions. These depthwise convolutions are initialized with the spatial components $x_{o,r}$ and $y_{o,r}$. \courier{Tucker} can be implemented with a pointwise convolution, followed by one regular 2D convolution. This second convolution is initialized with the core tensor $V_o$. In this network, the only trainable parameters include the new spectral components in the pointwise convolutions and the final classification layer. Unless specified, we use a decomposition rank of $R=2$. An illustration of all the different approaches can be found in figure \ref{fig:model_architecture}.
\subsubsection{Implementation details}
We use a batch size of 128 and a train-test split of 50\%-50\%. We do not use a validation due to the small size of the datasets. We use Adam with an exponentially decaying learning rate with $\gamma=0.95$ starting at 0.01, and cross-entropy loss over 100 epochs. We do not use weight decay, early stopping, or data augmentation. We repeat every experiment over 10 different seeds.


\begin{table}[t]
\caption{Mean test accuracy (with unbiased standard error of the mean) and number of trainable parameters of different model types for rank $R=2$ using the DenseNet121 backbone. Largest mean is in bold, and underlined when a Friedman test shows significant differences (at $p=0.05$) between the means of \courier{Scratch}, \courier{CP}, and \courier{Tucker}.}
\label{table:results_regular}
\centering
\begin{tabular}{l l llllcccc}
\toprule
\multirow{2}[2]{*}{Datasets} & \multicolumn{4}{c}{Accuracy (\%)} &\  & \multicolumn{4}{c}{\#Trainable parameters} \\ \cmidrule{2-5} \cmidrule{7-10}
& \multicolumn{1}{c}{\courier{Reduce}} & \multicolumn{1}{c}{\courier{Scratch}} & \multicolumn{1}{c}{\courier{CP}} & \multicolumn{1}{c}{\courier{Tucker}} & & \multicolumn{1}{c}{\courier{Reduce}} & \multicolumn{1}{c}{\courier{Scratch}} & \multicolumn{1}{c}{\courier{CP}} & \multicolumn{1}{c}{\courier{Tucker}} \\ \midrule
Botswana  & 89.7$\pm$1.2 & \textbf{96.9}$\pm$0.6 & 96.6$\pm$0.4 & 96.3$\pm$0.6 & & 20k & 456k & 20k & 20k \\

Indian Pines & 86.7$\pm$0.3 & 92.2$\pm$0.4 & \underline{\textbf{93.7}}$\pm$0.5 & 92.8$\pm$0.5 & & 27k & 629k & 27k & 27k \\

\ \ \ Kennedy & \multirow{2}{*}{94.4$\pm$0.4} & \multirow{2}{*}{95.3$\pm$0.9} & \multirow{2}{*}{\underline{\textbf{97.6}}$\pm$0.4}& \multirow{2}{*}{97.3$\pm$0.5} &\multirow{2}{*}{ } & \multirow{2}{*}{24k} & \multirow{2}{*}{553k} & \multirow{2}{*}{24k} & \multirow{2}{*}{24k} \\
Space Center & & & & & & & &  \\

\midrule
Avocado & 78.4$\pm$0.7 & \textbf{80.7}$\pm$1.2& 79.7$\pm$0.9& 78.6$\pm$1.0& & 34k & 708k & 34k & 34k \\
Grape leaves & 65.5$\pm$1.6 & 62.3$\pm$1.8 & 69.9$\pm$2.0 & \underline{\textbf{70.9}}$\pm$2.3 & & 41k & 624k & 41k & 41k \\
\bottomrule
\end{tabular}
\end{table}

\subsection{Results}
\subsubsection{Question 1: decomposed filters versus other methods}
\begin{figure}
    \centering
    \includegraphics{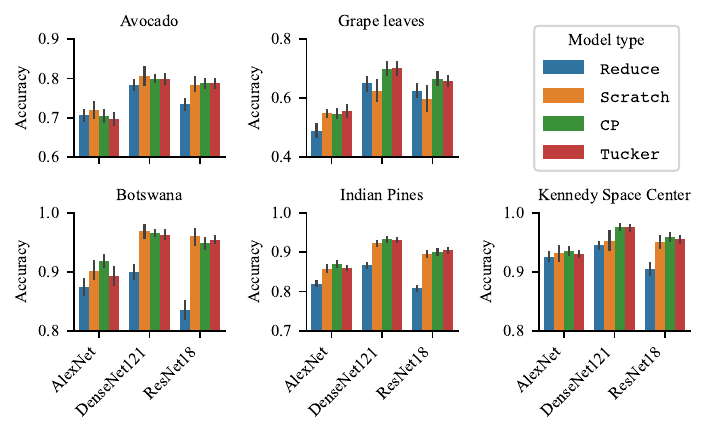}
    \caption{Final accuracy of different hyperspectral transfer learning methods with different backbones.}
    \label{fig:final_accuracy}
\end{figure}

Detailed results for the DenseNet121 backbone can be found in table \ref{table:results_regular}, results for all backbones are visualized in figure \ref{fig:final_accuracy}. On average, \courier{CP} and \courier{Tucker} are more accurate than \courier{Reduce} with roughly the same number of trainable parameters.
\courier{Scratch}, which uses many more trainable parameters, outperforms \courier{CP} and \courier{Tucker} in some cases, but the differences are small and in the cases where \courier{CP} and \courier{Tucker} do perform better the margin is more significant. We see the biggest difference for `Grape Leaves': \courier{Scratch} has the worst accuracy, even \courier{Reduce} outperforms it. This dataset is much more complex and diverse than the other datasets (visualized in figure \ref{fig:images}): the leaves vary in size and shape, and the diseases are present in distinct spatial and spectral patterns \cite{ryckewaert2023hyperspectral}.
\begin{figure}
    \centering
    \includegraphics[width=\linewidth]{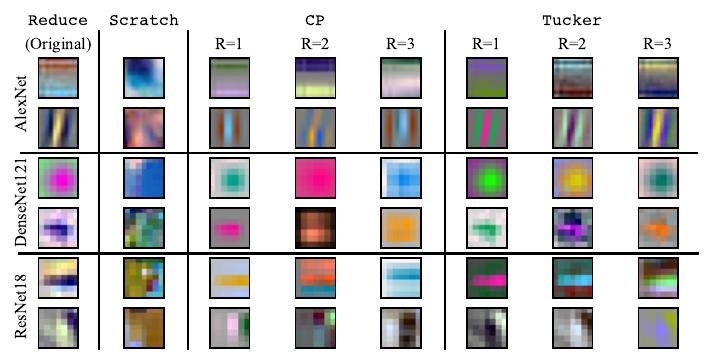}
    \caption{RGB-extracts of filters learned when using `Grape Leaves' for training. \courier{Reduce} uses the same filters as the original model. \courier{Scratch}'s filters are learned from scratch and hence do not resemble the original filters. \courier{CP} and \courier{Tucker} have the same spatial patterns as the original filter, but specialize the colors to the hyperspectral dataset.}
    \label{fig:plotted_filters}
\end{figure}
 \begin{figure}[b]
    \centering
    \includegraphics{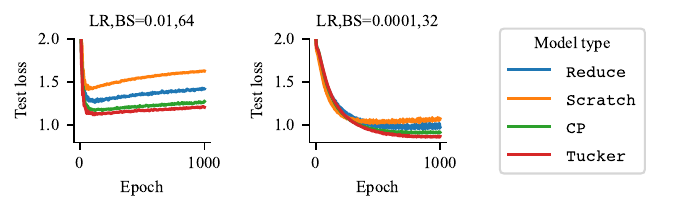}
    \caption{Two example loss curves over `Grape leaves' for a learning rate of 0.01 and batch size 64, and learning rate 0.0001 and batch size 32.}
    \label{fig:overfitting}
\end{figure}

For a more qualitative comparison, we visualize the learned hyperspectral filters by average pooling over the channel dimension (visualized in figure \ref{fig:plotted_filters}). \courier{Scratch}'s filters seem more random, showing that it might have overfit. For \courier{Tucker} we see that the original spatial patterns of the filter have been preserved almost perfectly. With \courier{CP} on the other hand, we do see some variation. As this decomposition type works with additive components, it is possible that a shape is split over several `subfilters', and different colors were learned for each subfilter. This may give the model some ability to learn new spatial patterns, particularly with higher ranks. 

A motivation for our method was the power to freeze parts of the backbone model, as retraining backbones on small datasets can cause overfitting. We now investigate whether \courier{CP} and \courier{Tucker} are actually more robust to overfitting than the other methods by intentionally over-optimizing the models. We repeat the experiment on the most complex dataset (`Grape leaves') with the most accurate backbone (Densenet121) using a variety of learning rates (0.0001, 0.001, 0.01) and batch sizes (4,8,16,32,64,128) and train for 1000 epochs over 10 seeds and record test loss. 
A selection of the loss curves are plotted in figure \ref{fig:overfitting}, the remaining loss curves can be found in the supplementary material. We see that for the first curve, \courier{Scratch} overfits faster than the other methods. With the more practical optimization scenario of the second curve, the overfitting is less drastic but \courier{Scratch} does seem to start overfitting slowly towards the end. The rest of the loss curves in the supplementary material  
all show the same pattern: if overfitting does happen, \courier{Scratch} starts doing so the earliest and fastest. 

\subsubsection{Question 2: choice of backbone}
We now compare accuracy over multiple backbones, visualized in figure \ref{fig:final_accuracy}. The choice of backbone does have significant impact on the overall accuracy, but this is in line with similar works \cite{varga2021measuring}. The conclusions from question 1 remain the same: \courier{CP} and \courier{Tucker} perform comparably, providing the overall highest accuracy. \courier{Scratch} is slightly less accurate: its lower mean accuracy can be attributed to the considerable gap on more complex datasets like `Grape leaves'. 
\subsubsection{Question 3: impact of decomposition choice}
Although CP and Tucker decompositions work in different ways, figure \ref{fig:final_accuracy} shows that they typically achieve similar accuracy in our models. The visualizations of learned filters in \ref{fig:plotted_filters} do show some differences between the methods, but it is difficult to determine how this affects performance. 
\begin{figure}
    \begin{subfigure}{1.7in}
        \centering
        \includegraphics{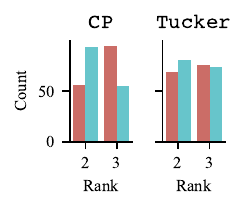} 
        \caption{}
        \label{fig:rank_count}
    \end{subfigure}
    ~ 
    \begin{subfigure}{2.5in}
        \centering
        \includegraphics{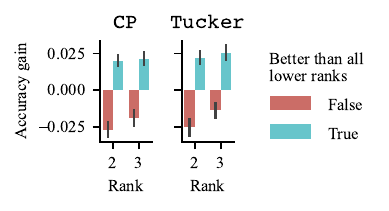} 
        \caption{}
        \label{fig:rank_error}
    \end{subfigure}
    \caption{(a) Number of experiments where a higher rank results in a higher accuracy. (b) Gain in accuracy when using a higher rank.}
    \label{fig:rank}
\end{figure}
We now investigate the impact of decomposition rank. In figure \ref{fig:rank_count} we visualize how many times a higher rank results in a higher score for the same experiment. In other words, we compare the accuracy of rank $R=2$ to $R=1$, and rank $R=3$ to both $R=2$ and $R=1$. For both decompositions, a rank of $R=2$ improves accuracy more often than not, but increasing the rank to $R=3$ does not improve it further generally. When considering the actual difference in accuracy in figure \ref{fig:rank_error}, we see that the acquired gain in accuracy when using a higher rank is not necessarily higher than the loss in accuracy when a lower rank performs better. In conclusion: rank does have impact, but needs to be chosen through hyperparameter search as a higher rank does not always improve accuracy. 

\section{Conclusion}
In this paper we addressed the incompatibility between hyperspectral images and RGB backbone models. We introduced a general approach that uses tensor decompositions of pretrained backbone filters to create new, partially trainable layers which preserve the backbone and specialize to the hyperspectral domain simultaneously. Our experiments show that our method is significantly more accurate than methods with similar complexity, and competitive to methods with higher complexity. Qualitative analysis revealed that the partially trainable layers can preserve the original spatial information of the filter, and that they are more robust to overfitting. 

\subsubsection{Limitations and future work}
In accordance with other works, we have also noticed that tensor decomposed layers can be more vulnerable to initialization methods and optimization strategies \cite{lebedev2015speeding}. An optimization strategy tailored to tensor decomposed layers could aid in making our approach more accurate and robust for larger neural networks \cite{phan2020stable,yin2021towards}. 

Finally, we already mentioned some problems with retraining the first layer of a backbone model: it can cause overfitting, the gradient can be unstable in early layers, and backpropagating through all layers can be computationally intensive. While our method combats the first two problems, it is still necessary to backpropagate through the entire network. Although it might not be possible to successfully apply 3-channel backbones to hyperspectral images without any specialization in the first layer, a method that can do so without needing to backpropagate through all layers could be a next step in creating universal backbones for hyperspectral imaging.




%
%
\bibliographystyle{splncs04}
\bibliography{main}
\clearpage
\appendix
\section{Loss curves}\label{appendix:overfitting}
\begin{figure}
    \centering
    \includegraphics{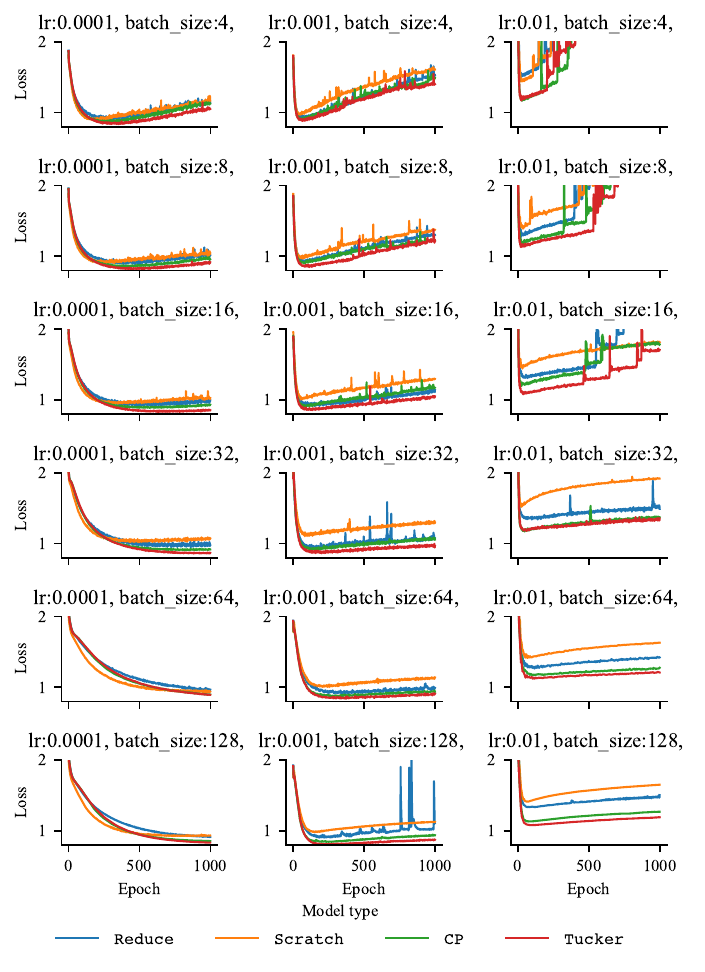}
    \caption{Loss curves over several combinations of learning rates and batch sizes on the `Grape leaves' dataset.}
    \label{fig:overfitting_large}
\end{figure}
\newpage

\section{Detailed results}
\begin{table}
\caption{Mean test accuracy (with unbiased standard error of the mean) and number of trainable parameters of different model types for rank $R=2$ using the AlexNet backbone. Largest mean is in bold, and underlined when a Friedman test shows significant differences (at $p=0.05$) between the means of \courier{Scratch}, \courier{CP}, and \courier{Tucker}.}
\label{table:results_alexnet}
\centering
\begin{tabular}{l l llllcccc}
\toprule
\multirow{2}[2]{*}{Datasets} & \multicolumn{4}{c}{Accuracy (\%)} &\  & \multicolumn{4}{c}{\#Trainable parameters} \\ \cmidrule{2-5} \cmidrule{7-10}
& \multicolumn{1}{c}{\courier{Reduce}} & \multicolumn{1}{c}{\courier{Scratch}} & \multicolumn{1}{c}{\courier{CP}} & \multicolumn{1}{c}{\courier{Tucker}} & & \multicolumn{1}{c}{\courier{Reduce}} & \multicolumn{1}{c}{\courier{Scratch}} & \multicolumn{1}{c}{\courier{CP}} & \multicolumn{1}{c}{\courier{Tucker}} \\ \midrule
Botswana  & 88.5$\pm$1.1 & 90.1$\pm$0.8 & \textbf{92.8}$\pm$1.0 & 90.5$\pm$1.3 & & 21k & 1125k & 21k & 21k \\

Indian Pines & 82.0$\pm$0.6 & 85.8$\pm$0.5 & \underline{\textbf{87.8}}$\pm$0.7 & 86.1$\pm$0.5 & & 28k & 1551k & 28k & 28k \\

\ \ \ Kennedy & \multirow{2}{*}{92.7$\pm$0.7} & \multirow{2}{*}{93.2$\pm$0.7} & \multirow{2}{*}{\textbf{94.0}$\pm$0.4}& \multirow{2}{*}{92.9$\pm$0.5} &\multirow{2}{*}{ } & \multirow{2}{*}{25k} & \multirow{2}{*}{1365k} & \multirow{2}{*}{25k} & \multirow{2}{*}{25k} \\
Space Center & & & & & & & &  \\

\midrule
Avocado & 70.3$\pm$1.2 & \textbf{72.0}$\pm$1.2 & 71.4$\pm$1.5 & 69.3$\pm$1.6 & & 37k & 1743k & 37k & 37k \\
Grape leaves & 46.2$\pm$2.0 & 54.8$\pm$0.7 & \textbf{54.9}$\pm$1.6 & 54.6$\pm$2.0 & & 49k & 1526k & 49k & 49k \\
\bottomrule
\end{tabular}
\end{table}

\begin{table}
\caption{Mean test accuracy (with unbiased standard error of the mean) and number of trainable parameters of different model types for rank $R=2$ using the ResNet18 backbone. Largest mean is in bold, and underlined when a Friedman test shows significant differences (at $p=0.05$) between the means of \courier{Scratch}, \courier{CP}, and \courier{Tucker}.}
\label{table:results_resnet18}
\centering
\begin{tabular}{l l llllcccc}
\toprule
\multirow{2}[2]{*}{Datasets} & \multicolumn{4}{c}{Accuracy (\%)} &\  & \multicolumn{4}{c}{\#Trainable parameters} \\ \cmidrule{2-5} \cmidrule{7-10}
& \multicolumn{1}{c}{\courier{Reduce}} & \multicolumn{1}{c}{\courier{Scratch}} & \multicolumn{1}{c}{\courier{CP}} & \multicolumn{1}{c}{\courier{Tucker}} & & \multicolumn{1}{c}{\courier{Reduce}} & \multicolumn{1}{c}{\courier{Scratch}} & \multicolumn{1}{c}{\courier{CP}} & \multicolumn{1}{c}{\courier{Tucker}} \\ \midrule
Botswana  & 83.8$\pm$1.4 & \textbf{96.0}$\pm$0.7 & 95.7$\pm$0.6 & 95.4$\pm$0.6 & & 20k & 456k & 20k & 20k \\

Indian Pines & 81.1$\pm$0.7 & 89.5$\pm$0.4 & 90.1$\pm$0.7 & \textbf{90.7}$\pm$0.7 & & 27k & 629k & 27k & 27k \\

\ \ \ Kennedy & \multirow{2}{*}{90.6$\pm$1.0} & \multirow{2}{*}{95.0$\pm$0.5} & \multirow{2}{*}{\underline{\textbf{96.6}}$\pm$0.4}& \multirow{2}{*}{95.9$\pm$0.7} &\multirow{2}{*}{ } & \multirow{2}{*}{24k} & \multirow{2}{*}{553k} & \multirow{2}{*}{24k} & \multirow{2}{*}{24k} \\
Space Center & & & & & & & &  \\

\midrule
Avocado & 74.2$\pm$1.6 & 78.4$\pm$0.9 & 78.1$\pm$1.2 & \textbf{78.6}$\pm$1.1 & & 34k & 708k & 34k & 34k \\
Grape leaves & 61.3$\pm$1.8 & 59.8$\pm$2.2 & \underline{\textbf{66.1}}$\pm$2.1 & 64.0$\pm$1.3 & & 41k & 624k & 41k & 41k \\
\bottomrule
\end{tabular}
\end{table}
\end{document}